\documentclass[conference]{IEEEtran}

\makeatletter
\def\ps@IEEEtitlepagestyle{%
  \def\@oddfoot{\mycopyrightnotice}%
  \def\@evenfoot{}%
}
\def\mycopyrightnotice{%
  {\footnotesize 978-1-6654-7095-7/22/\$31.00~\copyright~2022 IEEE\hfill}
  \gdef\mycopyrightnotice{}
}

\usepackage{blindtext}
\usepackage{eso-pic}
\IEEEoverridecommandlockouts
\usepackage{cite}
\usepackage{amsmath,amssymb,amsfonts}
\usepackage{algorithmic}
\usepackage{graphicx}
\usepackage{textcomp}
\usepackage{xcolor}
\def\BibTeX{{\rm B\kern-.05em{\sc i\kern-.025em b}\kern-.08em
    T\kern-.1667em\lower.7ex\hbox{E}\kern-.125emX}}
    
\usepackage{eso-pic}
\newcommand\AtPageUpperMyright[1]{\AtPageUpperLeft{%
 \put(\LenToUnit{0.17\paperwidth},\LenToUnit{-2cm}){%
     \parbox{0.9\textwidth}{\raggedleft\fontsize{8}{11}\selectfont #1}}%
 }}%
\newcommand{\conf}[1]{%
\AddToShipoutPictureBG*{%
\AtPageUpperMyright{#1}
}
}    
    
\usepackage{booktabs}
\usepackage{listings}
\usepackage{scalerel}
\usepackage[binary-units]{siunitx}
\usepackage{subcaption}
\usepackage{tablefootnote}
\usepackage[textsize=tiny]{todonotes}
\usepackage{tikz, pgfplots}
  \usetikzlibrary{svg.path}

\makeatletter
\let\NAT@parse\undefined
\makeatother
\usepackage{hyperref}

\newcommand\httpsurl[1]{%
  \href{https://#1}{\nolinkurl{#1}}%
}

\definecolor{rwth-blue}{cmyk}{1, 0.5, 0, 0}
\definecolor{rwth-black}{cmyk}{0, 0, 0, 1}
\definecolor{rwth-magenta}{cmyk}{0, 1, 0.25, 0}
\definecolor{rwth-yellow}{cmyk}{0, 0, 1, 0}
\definecolor{rwth-petrol}{cmyk}{1, 0.3, 0.5, 0.3}
\definecolor{rwth-turquoise}{cmyk}{1, 0, 0.4, 0}
\definecolor{rwth-green}{cmyk}{0.7, 0, 1, 0}
\definecolor{rwth-maygreen}{cmyk}{0.35, 0, 1, 0}
\definecolor{rwth-orange}{cmyk}{0, 0.4, 1, 0}
\definecolor{rwth-red}{cmyk}{0.15, 1, 1, 0}
\definecolor{rwth-bordeaux}{cmyk}{0.25, 1, 0.7, 0.2}
\definecolor{rwth-violet}{cmyk}{0.7, 1, 0.35, 0.15}
\definecolor{rwth-purple}{cmyk}{0.6, 0.6, 0, 0}

\definecolor{orcidlogocol}{HTML}{A6CE39}
\tikzset{
  orcidlogo/.pic={
    \fill[orcidlogocol] svg{M256,128c0,70.7-57.3,128-128,128C57.3,256,0,198.7,0,128C0,57.3,57.3,0,128,0C198.7,0,256,57.3,256,128z};
    \fill[white] svg{M86.3,186.2H70.9V79.1h15.4v48.4V186.2z}
                 svg{M108.9,79.1h41.6c39.6,0,57,28.3,57,53.6c0,27.5-21.5,53.6-56.8,53.6h-41.8V79.1z M124.3,172.4h24.5c34.9,0,42.9-26.5,42.9-39.7c0-21.5-13.7-39.7-43.7-39.7h-23.7V172.4z}
                 svg{M88.7,56.8c0,5.5-4.5,10.1-10.1,10.1c-5.6,0-10.1-4.6-10.1-10.1c0-5.6,4.5-10.1,10.1-10.1C84.2,46.7,88.7,51.3,88.7,56.8z};
  }
}
\newcommand\orcidicon[1]{\href{https://orcid.org/#1}{\mbox{\scalerel*{
\begin{tikzpicture}[yscale=-1,transform shape]
\pic{orcidlogo};
\end{tikzpicture}
}{|}}}}

\newcommand{\fig}[1]{Fig.~\ref{#1}}

\newcommand{\tab}[1]{Table~\ref{#1}}

\newcommand{\sect}[1]{Section~\ref{#1}}

\begin{document}

\bstctlcite{IEEEexample:BSTcontrol}

\title{\vspace*{1cm} Enabling Connectivity for Automated Mobility: \\
  A Novel MQTT-based Interface Evaluated in a 5G
  Case Study on Edge-Cloud Lidar Object Detection
\thanks{*These authors contributed equally to this work.}%
}

\author{\IEEEauthorblockN{Lennart Reiher*\textsuperscript{\orcidicon{0000-0002-7309-164X}}}
\IEEEauthorblockA{\textit{Institute for Automotive Engineering} \\
\textit{RWTH Aachen University}\\
Aachen, Germany \\
lennart.reiher@ika.rwth-aachen.de}
\and
\IEEEauthorblockN{Bastian Lampe*\textsuperscript{\orcidicon{0000-0002-4414-6947}}}
\IEEEauthorblockA{\textit{Institute for Automotive Engineering} \\
\textit{RWTH Aachen University}\\
Aachen, Germany \\
bastian.lampe@ika.rwth-aachen.de}
\and
\IEEEauthorblockN{Timo Woopen\textsuperscript{\orcidicon{0000-0002-7177-181X}}}
\IEEEauthorblockA{\textit{Institute for Automotive Engineering} \\
\textit{RWTH Aachen University}\\
Aachen, Germany \\
timo.woopen@ika.rwth-aachen.de}
\and
\IEEEauthorblockN{Raphael van Kempen\textsuperscript{\orcidicon{0000-0001-5017-7494}}}
\IEEEauthorblockA{\textit{Institute for Automotive Engineering} \\
\textit{RWTH Aachen University}\\
Aachen, Germany \\
raphael.vankempen@ika.rwth-aachen.de}
\and
\IEEEauthorblockN{Till Beemelmanns\textsuperscript{\orcidicon{0000-0002-2129-4082}}}
\IEEEauthorblockA{\textit{Institute for Automotive Engineering} \\
\textit{RWTH Aachen University}\\
Aachen, Germany \\
till.beemelmanns@ika.rwth-aachen.de}
\and
\IEEEauthorblockN{Lutz Eckstein}
\IEEEauthorblockA{\textit{Institute for Automotive Engineering} \\
\textit{RWTH Aachen University}\\
Aachen, Germany \\
lutz.eckstein@ika.rwth-aachen.de}
}

\maketitle
\conf{\textit{  Proc. of the International Conference on Electrical, Computer, Communications and Mechatronics Engineering  (ICECCME) \\ 
16-18 November 2022, Maldives}}
\begin{abstract}

Enabling secure and reliable high-bandwidth low-latency connectivity between automated vehicles and external servers, intelligent infrastructure, and other road users is a central step in making fully automated driving possible.
The availability of data interfaces, which allow this kind of connectivity, has the potential to distinguish artificial agents' capabilities in connected, cooperative, and automated mobility systems from the capabilities of human operators, who do not possess such interfaces.
Connected agents can for example share data to build collective environment models, plan collective behavior, and learn collectively from the shared data that is centrally combined.
This paper presents multiple solutions that allow connected entities to exchange data. In particular, we propose a new universal communication interface which uses the Message Queuing Telemetry Transport~(MQTT) protocol to connect agents running the Robot Operating System~(ROS). Our work integrates methods to assess the connection quality in the form of various key performance indicators in real-time. 
We compare a variety of approaches that provide the connectivity necessary for the exemplary use case of edge-cloud lidar object detection in a 5G network. We show that the mean latency between the availability of vehicle-based sensor measurements and the reception of a corresponding object list from the edge-cloud is below~\SI{87}{\milli\second}. All implemented solutions are made open-source and free to use. Source code is available at \httpsurl{github.com/ika-rwth-aachen/ros-v2x-benchmarking-suite}.

\end{abstract}



\section{Introduction}
\label{sec:introduction}

\begin{figure}[!ht]
  \centering
  \vspace{0.5\baselineskip}
  \includegraphics[width=0.75\columnwidth, trim=9.3cm 4.1cm 9.5cm 3.0cm, clip]{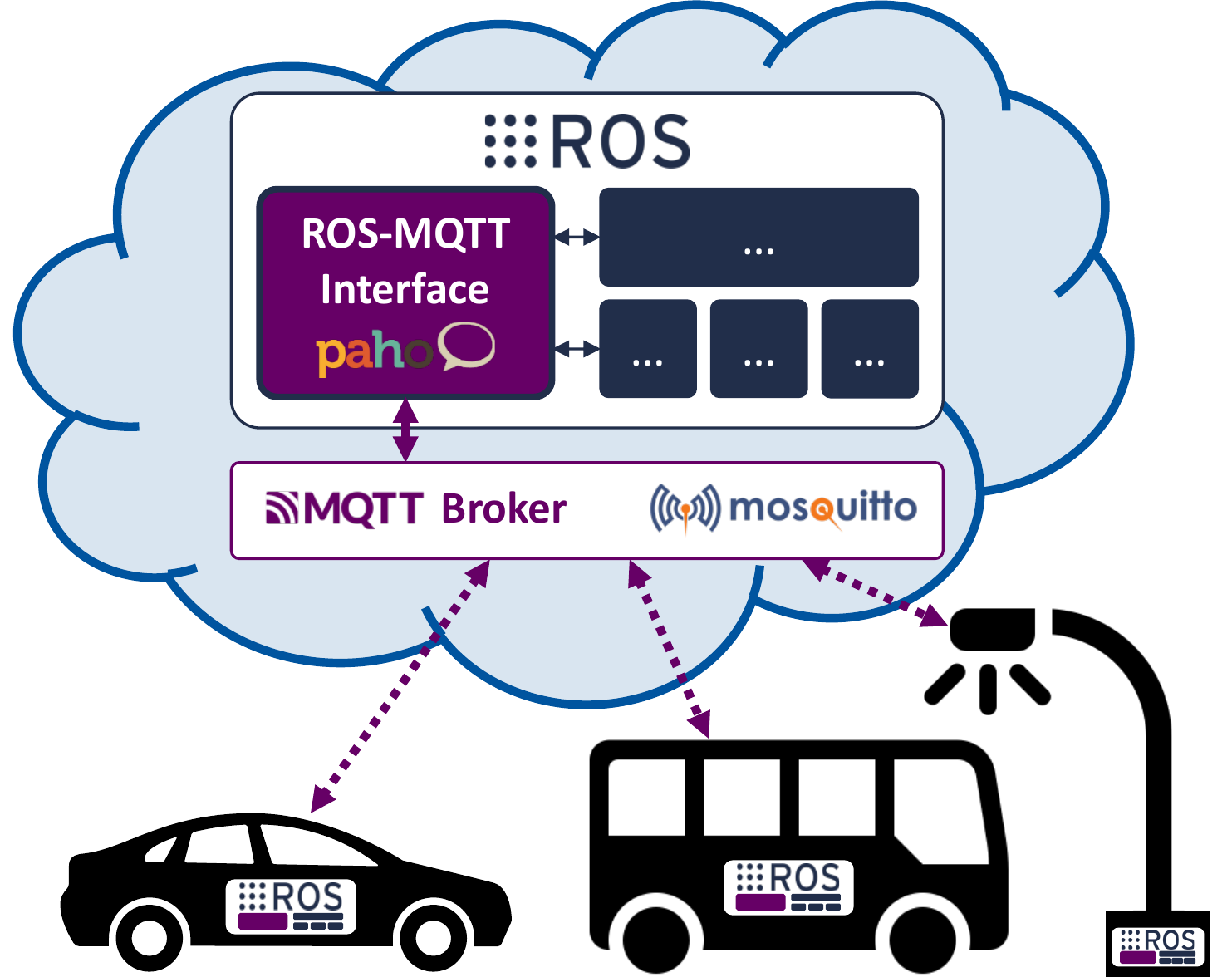} 
  \caption{Evaluated approach to cloud-assisted connected driving functions: intelligent agents (e.g., automated vehicles, intelligent infrastructure) are running on ROS; a dedicated edge-cloud or cloud server is also running ROS to realize the collective potential of data shared by the connected agents; data transmission is based on MQTT, with a newly developed ROS-MQTT interface handling the communication between all connected entitites, including the server.}
  \label{fig:cloud-architecture}
\end{figure}

Connectivity allows existing automated driving functions to be substantially improved and entirely new functions to be conceived~\cite{Woopen_UNICARagil_2018_ACK, Lampe_Collective_2019_ACK}. Cloud-based data processing makes advanced hardware, capable of running large and powerful models, available to connected vehicles or other connected entities. 
In addition, data sharing enables the combination of data from multiple road users, which allows solving challenges such as occlusions in sensor data or the prediction of road users' behavior.
There is a trend towards extremely large models in deep learning. Larger models usually show better performance~\cite{BubeckSellke_UniversalLawRobustness_2021,SevillaEtAl_ComputeTrendsThree_2022, Zhai_Scaling_2021_ARXIV, Kaplan_Scaling_2020_ARXIV} but also require more energy during training and inference~\cite{Martin_Energy_2019_JPDC}. This makes offloading functions for automated driving more and more relevant, because vehicles have relatively limited computational resources and a limited supply of energy.
Since a safe operation of an automated vehicle relies on models of high performance, there exists a substantial incentive to offload functions to more powerful servers with more powerful models, and to overcome the challenges associated with this task.
In the past, offloading automated vehicle functions to external servers has been challenging, especially due to the limited bandwidth and large latencies in cellular networks. With the introduction of 5G, remote processing becomes more feasible.
While earlier concepts based on 4G usually processed small data such as object lists remotely~\cite{Buchholz_MEC_2021_ITSM}, 5G also allows to process raw sensor data~\cite{Garcia_V2X_2021_CST} such as images and point clouds. Such large unstructured types of data are especially suited to be processed by large deep learning models.

A prerequisite for exchanging data between connected entitites is to have a flexible communication interface, which is integrated into both the software architecture of connected agents and that of supporting infrastructure such as cloud servers. Such an interface is developed, evaluated, and released as part of this paper.

\subsection{Contribution}
\label{ssec:contribution}

The contribution of this work is three-fold:

\begin{itemize}
  \item We motivate, explain, and implement a novel MQTT-based communication interface for robotic entities, and apply it in the context of connected automated driving.
  \item We present an in-depth analysis of using the new interface and presented alternatives for a promising use case in automated driving, namely edge-cloud lidar object detection. We detail the various solutions and their associated advantages and disadvantages. This can help other researchers to choose and build upon the best solution for their use case.
  \item We release the newly implemented communication interface as free and open-source code, such that it gets substantially easier for anyone to develop connected robotics systems capable of exchanging and processing large data at low latencies. All additional tooling and data for the use case analysis is also made available. This allows others to analyze the presented use case in their own network and hopefully facilitates other researchers' work.
\end{itemize}

\section{Related Work}
\label{sec:related-work}

Before explaining the approach developed as part of this work, we will first give a brief overview of use cases in the context of automated driving that could benefit from our developed solutions. Then, we provide an overview of publications that present system architectures into which our solutions could be incorporated. Next, we present studies that compare different communication protocols, which compete with the ones presented here. Last, we summarize studies that are similar to the one at hand and to which the performance indicators presented in this work can be compared.

\subsection{Connected Driving Functions}

Connected driving functions can be structured into cooperative, collective, and supportive functions. These categories differ with respect to whether the data is processed centrally vs.\ decentrally, whether data from multiple entities is combined, and whether the benefits of the data exchange are symmetric. 

In cooperative functions, data is shared among connected entities and processed decentrally in these entities. Connected agents may use this data exchange to cooperate. A prominent example is cooperative perception. A description of how perceived objects can be shared with other vehicles such that they can incorporate them into their environment model can be found in~\cite{Yoon_Cooperative_2021_TITS} and~\cite{Rauch_V2X_2011_IV}.

In collective functions, data is collected from multiple connected entities and processed centrally. The results may be transmitted to and used by the collective of connected entities. Examples include collective environment modeling and collective learning. \cite{Lampe_Collective_2019_ACK} describes how environment models of connected automated vehicles can be combined in an edge-cloud server. The collective environment model is then shared with relevant connected entities. \cite{Karpathy_Autolabeling_2021_CVPR} describes how data from a fleet of vehicles can be collected to create new training samples for improving existing functions of connected vehicles. The authors of~\cite{Kloeker_Traffic_2020_ITSC} describe how data from connected infrastructure stations can be used for the development and validation of automated driving functions.

For supportive functions, a connected entity provides data or processing power as a service to other individual connected entities. Examples include the provision of traffic light states, traffic monitoring and function offloading. In~\cite{Guo_Signals_2019_TR}, we find an overview of how the transmission of traffic light states to automated vehicles can improve behavior planning. In~\cite{Buchholz_MEC_2021_ITSM}, infrastructure sensors are used to provide detected objects and predictions to automated vehicles. The authors of~\cite{Kumar_Cloud_2012_MCC} describe how automated vehicles can offload parts of the environment perception and planning to a cloud server.

\subsection{Connected Driving System Architectures}

Another useful categorization concerns the communication architecture used for connected driving. The architecture particularly follows the latency requirements of the respective use cases. Among other aspects, the limited speed of light requires that the stricter the latency requirements, the closer the data processing entity needs to be to the data source. A use case such as collective learning does not rely on very low latencies, so a single centralized cloud architecture may handle numerous connected entities in a large area. For latency-sensitive use cases such as function offloading in the form of lidar object detection, multi-access edge computing~(MEC) may be preferred, because the decreased distance between connected agents and the data processing server leads to lower latencies. A generalized concept of MEC and cloud computing is called fog computing, where there exists a mix of data processing entities responsible for different use cases depending on the respective latency requirement. In~\cite{Mao_MEC_2017_CST}, we find a general overview of research regarding MEC and fog computing applications, whereas~\cite{Zhang_IoV_2020_IEEE} specifically focuses on aspects regarding automated driving.

\subsection{Communication Protocols}

The protocol stack used in connected driving use cases consists of multiple layers, e.g., as described by the OSI-model. While technologies such as 5G, ITS-G5, and DSRC relate to the lower layers in the protocol stack, applications may choose among different options on the application, presentation, and session layer. A survey conducted by~\cite{Dizdarevic_Survey_2018_ACM} analyzes different protocols such as MQTT, AMQP, DDS, HTTP, and CoAP in the context of the Internet of Things~(IoT) in a fog computing architecture. They conclude that MQTT and RESTful HTTP are the most mature choices to consider. MQTT is more similar to ROS-based systems in that it supports a publish-subscribe type of interaction. It is also better suited for systems in which the communication and battery consumption are restrained. 

\subsection{Reference Solutions and 5G Studies}

There exists one reference implementation of a ROS- and MQTT-based communication interface published at~\cite{mqtt_bridge} to which our developed MQTT-based communication interface can be compared. \cite{Mukhandi_Secure_2019_SII}~shows how this reference solution can be used to securely exchange data in a network of ROS-based robots. The authors of~\cite{Mokhtarian_Cloud_2021_CTS} evaluate a ROS-MQTT interface against a FastRTPS-based approach. There also exist implementations~\cite{udp_bridge, fkie_multimaster, Tiderko_Multimaster_2016_ROS, ros2} that may represent viable alternatives to the solutions described in this paper but exceed the scope of this work. Especially ROS~2~\cite{ros2} is promising as soon as it has matured sufficiently to be used in a wide variety of applications.

Some real-world studies evaluating automated driving use cases with respect to latencies in 5G networks have already been conducted.
The authors of~\cite{Kutila_5G_2021_IV} report latencies between \SI{43}{\milli\second} and \SI{130}{\milli\second} depending on the chosen experimental setup. They analyze the use case of sending data from a road side unit to an automated vehicle via a 5G NSA network.
The authors of~\cite{Hetzer_5G_2021_WCN} report mean round trip times of \SI{8.1}{\milli\second} as measured with the \lstinline{ping} tool and a mean application-specific round trip time of \SI{16.8}{\milli\second} for the use case of anticipated cooperative collision avoidance. 
In~\cite{Ficzere_5G_2021_IM}, we find the evaluation of latencies in a 5G NSA network using a moving vehicle. They report mean latencies between \SI{9.2}{\milli\second} and \SI{18.6}{\milli\second} depending on message size.

There have also been studies of using a 5G network indoors in an Industry~4.0 setting. The authors of~\cite{Voigtlaender_5GROS_2017_ISCSIC} report one-way latencies of around \SI{2}{\milli\second} when controlling a robot arm. The work of~\cite{Ansari_5G_2022_ELE} reports median latencies of below \SI{0.8}{\milli\second} in uplink and \SIrange{0.8}{0.9}{\milli\second} in downlink. The authors of~\cite{Rischke_5G_2021_IEEE} report latencies below \SI{10}{\milli\second} in their indoor testbed setup. 

All studies have in common that their results are very difficult to reproduce, since the transmitted data, the experiment software and the evaluation software are not made publicly available. Since transmission delays are very sensitive to the experimental setup, reported network performance indicators are difficult to compare. Our work therefore aims to provide an easy to use implementation that can be used for various use cases and that makes results easy to compare.

\section{Methodology}
\label{sec:methodology}

Data exchange among connected entities or to a cloud system is facilitated by a tight integration into the software architectures of all involved clients. In the domain of robotics, the Robot Operating System~(ROS) is by far the most popular framework and collection of open-source software components. It is therefore advantageous if a data interface for the communication between ROS-based entities is easy to integrate with ROS.

Our proposed interface connects agents running ROS via the MQTT protocol. In the following, we first motivate the overall architecture and then detail the capabilities of our open-sourced ROS-MQTT interface.

\subsection{Communication Interface for Connected Driving}
\label{ssec:cloud-architecture-for-connected-driving}

Apart from performance requirements and its integration ability with ROS, a communication protocol for connected entities should fulfil additional demands that arise in the context of connected driving.

Single points of failure should be avoided. If the connection between clients is impaired, entities should be able to remain operational with respect to non-connected functions. This suggests to separate local coordinator functions (e.g., a ROS master process or another kind of orchestrator~\cite{Mokhtarian_ASOA_2020_ACK}) from communication-coordinating functions. In general, the system architecture should be able to handle transmission disturbances as well as entire interruptions of communication with stable and defined behavior, ideally reconnecting automatically.

The security and integrity of transmitted data has to be ensured at all times, if safety-relevant data is concerned, as is usually the case with automated driving functions. Data security is also required to protect sensible data under privacy concerns.

Derived from the requirements mentioned above, we propose using a universal data interface based on MQTT for connecting agents running ROS. As a publish-subscribe network protocol, MQTT is similar to the communication pattern used within ROS itself. MQTT usually runs on the TCP transport layer. Communication among many MQTT clients is coordinated through an MQTT broker, which routes messages based on current message subscriptions to their designated destination. MQTT supports different Quality-of-Service~(QoS) levels, authentication, as well as the encryption of all transmitted data.

Our proposed ROS-based interface for connectivity among intelligent agents, enabled via MQTT, is illustrated in \fig{fig:cloud-architecture}. All involved entities, i.e., both intelligent clients such as automated vehicles and remote servers, run their functions on ROS. The server has access to data shared by the intelligent clients, enabling it to realize connected driving functions. Bi-directional data transmission between cloud and clients is relayed via the MQTT broker (\textit{Eclipse Mosquitto}~\cite{mosquitto} in our case), which also runs on the cloud node. The transfer between ROS and MQTT protocols is handled via a dedicated ROS-MQTT interface, which is running on all clients including the cloud. The approach is also equivalently applicable to other kinds of communication among participants, e.g., V2V in a connected driving setting. Note that all major components of this approach are open-source tools.

\subsection{ROS-MQTT Interface}
\label{ssec:ros-mqtt-interface}

Our ROS-MQTT interface is specifically implemented in consideration of the requirements described before. It is designed as a generic and use case-agnostic interface between ROS and MQTT. The interface is released as a ROS package called \lstinline{mqtt_client}\footnote{\httpsurl{github.com/ika-rwth-aachen/mqtt_client}} and is made open-source and free to use.

Although one such implementation already exists in the ROS ecosystem (packaged as \lstinline{mqtt_bridge}~\cite{mqtt_bridge}), it does not seem to be focused on performance as a ROS Python node. Our novel interface is instead implemented as a C++ ROS nodelet. Compared to Python, this not only improves general performance in theory, but also enables no-copy ROS message transfers to and from the interface.

Regarding universal application to arbitrary types of data, a challenge with C++ is its lack of runtime introspection capabilities, which the Python bridge fundamentally relies on. Through generic topic subscribers and publishers, our C++ interface nonetheless is universally compatible with arbitrary ROS message types, including custom ones. Any published ROS topic can simply be mapped to an MQTT topic and vice-versa. Note that the interface is primarily designed to exchange data with other instances of \lstinline{mqtt_client}. In principle however, it is also possible to communicate with MQTT clients not powered by ROS, e.g., microcontrollers or constrained IoT devices.

Under the hood, the \textit{Paho MQTT C++ Client}~\cite{paho_mqtt} library is used to connect to an MQTT broker. All configuration options provided by the library are exposed as ROS parameters: this includes authentication, encryption, QoS, buffering, and more. An exemplary configuration for a simple communication between a vehicle and a cloud is presented in~\fig{fig:mqtt-client-config}.

\begin{figure}[!t]
  \begin{subfigure}[b]{0.45\columnwidth}
    \begin{lstlisting}[basicstyle=\scriptsize\ttfamily]
broker:
  host: cloud
  user: admin
  pass: password
client:
  id: vehicle
bridge:
  ros2mqtt:
    - ros_topic: /ping
      mqtt_topic: ping
  mqtt2ros:
    - mqtt_topic: pong
      ros_topic: /pong
    \end{lstlisting}
    \caption{Vehicle configuration}
  \end{subfigure}
  \hfill
  \begin{subfigure}[b]{0.45\columnwidth}
    \begin{lstlisting}[basicstyle=\scriptsize\ttfamily]
broker:
  host: localhost
  user: admin
  pass: password
client:
  id: cloud
bridge:
  mqtt2ros:
    - mqtt_topic: ping
      ros_topic: /ping
  ros2mqtt:
    - ros_topic: /ping
      mqtt_topic: pong
    \end{lstlisting}
    \caption{Cloud configuration}
  \end{subfigure}
  \caption{Exemplary configuration of \lstinline{mqtt_client} for simple ping-pong communication from vehicle to cloud and back: ROS messages on topic \lstinline{/ping} on the vehicle are forwarded to MQTT topic \lstinline{ping}, which is re-published as \lstinline{/ping} in the cloud's ROS network. This message is instantly returned back to MQTT topic \lstinline{pong}, which at the end is re-published as \lstinline{/pong} on the vehicle's ROS network. The initial message can be of arbitrary ROS message type.}
  \label{fig:mqtt-client-config}
\end{figure}

\subsection{Use Case: 5G Edge-Cloud Lidar Object Detection}
\label{ssec:use-case-5g-edge-cloud-lidar-object-detection}

We choose 5G-enabled edge-cloud lidar object detection as the connected driving-related use case for the evaluation of our proposed data interface.

An automated vehicle running ROS is connected to a powerful edge-cloud server via 5G. The vehicle transmits point clouds to the server, where an artificial neural network processes the data and computes lists of objects detected in the point clouds. The object lists are transmitted back to the vehicle, where they could be incorporated into vehicle functions such as behavior planning. The analysis splits the total latency into all relevant parts, namely the propagation, computation, and communication latencies~\cite{Mao_MEC_2017_CST}, which are associated with the different sections of the data pipeline.

\section{Experimental Setup}
\label{sec:experimental-setup}

In order to evaluate our MQTT-based approach to connected driving -- as presented in \sect{sec:methodology} -- we run multiple experiments and measure relevant metrics as performance indicators. In the following, we present the experimental setup and benchmarking process.

\subsection{Benchmarking Methodology}
\label{ssec:benchmarking-methodology}

The main experimental setup involves the following steps:
\begin{enumerate}
  \item send a lidar point cloud from a 5G-connected vehicle to a cloud server;
  \item perform neural network-based lidar object detection in the cloud to infer a list of objects from the point cloud;
  \item send the list of detected objects back to the vehicle.
\end{enumerate}

\begin{figure*}[!t]
  \centering
  \includegraphics[width=\textwidth, trim=1.9cm 7.1cm 1.9cm 7.1cm, clip]{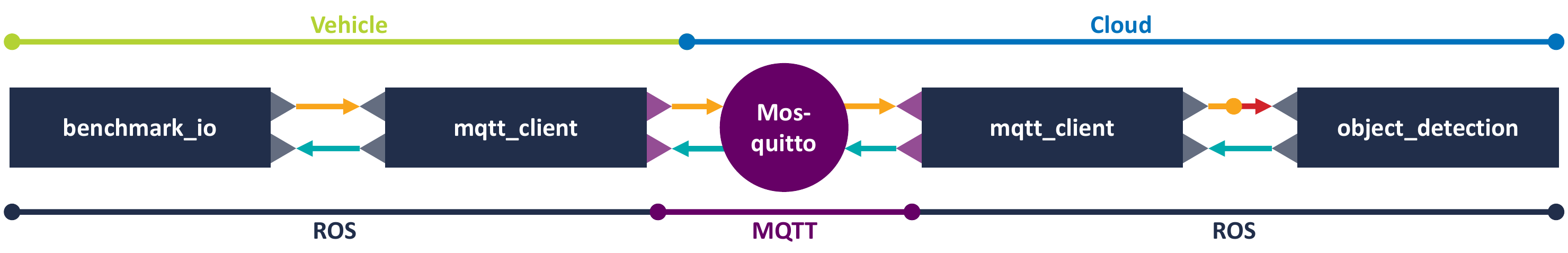} 
  \caption{\textcolor{rwth-orange}{Point clouds} are published on the vehicle and transmitted to the cloud via MQTT, using two instances of \lstinline{mqtt_client}. Before entering the object detection, the \textcolor{rwth-orange}{\lstinline{VelodyneScan} point cloud messages} are transformed to \textcolor{rwth-red}{\lstinline{PointCloud2} point cloud messages}. \textcolor{rwth-turquoise}{Object lists} are then computed from incoming point clouds and sent back to the vehicle. In- and out-timestamps are taken at each component, depicted as little triangles~\(\triangleright\)/\(\triangleleft\).}
  \label{fig:data-flow}
\end{figure*}

The dataflow described in the following is illustrated in \fig{fig:data-flow}. The starting point for each experiment is the availability of a ROS message containing a point cloud, as would be published by any lidar ROS driver. In our case, this means having a \lstinline{velodyne_msgs::VelodyneScan} ROS message available in memory, which is the output of one \SI{360}{\degree} scan of a \textit{Velodyne VLP-32C} lidar sensor. To ensure a fair comparison between multiple experiments, we work with a fixed point cloud. The point cloud used for the experiments contains a total of \num{49016} points, each including coordinates in 3D space as well as an intensity value. Unless stated otherwise, we publish this point cloud at \SI{10}{\hertz}.

The published point cloud ROS message is subscribed by an instance of \lstinline{mqtt_client}, which serializes the message and publishes it to an MQTT broker. As broker, we use the popular open-source \textit{Eclipse Mosquitto}~\cite{mosquitto}, which we run on the edge-cloud server. The broker's main task is to distribute data published by its clients to other clients subscribing to that particular data topic. In our case, the point cloud message is forwarded to a second instance of the \lstinline{mqtt_client}, this time running on the edge-cloud server. There, the message is converted back to its original ROS message format and then published on the ROS network.

On the edge-cloud server, we run a neural network-based object detection algorithm. It is implemented as a \textit{TensorFlow}~\cite{tensorflow2015-whitepaper} model, executed together with necessary preprocessing and postprocessing steps in a C++ ROS nodelet. Note that neither the prediction performance nor the runtime performance of the object detection algorithm are optimized or evaluated in this work. The object detection nodelet expects a \lstinline{sensor_msgs::PointCloud2} message as input, which is the most commonly used format for working with point clouds in ROS. In order to convert the transmitted \lstinline{VelodyneScan} message, we additionally run a transformation nodelet that is part of the official \textit{Velodyne} ROS driver packages~\cite{velodyne}. The output of the object detection nodelet is an object list containing object poses as detected in the point cloud. The inferred object list is published as a ROS message of custom message type.

In the same way as the point cloud, the object list is transmitted back to the vehicle via the two \lstinline{mqtt_client} instances running on the edge-cloud and the vehicle computer, respectively.

In order to measure relevant performance indicators for the process and its parts, we log in- and out-timestamps at every component that data passes through. These timestamps are aggregated on the vehicle computer to compute live metrics during the experiments. The timestamps taken on the edge-cloud's components are also transmitted to the vehicle via the MQTT interface. Note that while we set up the vehicle computer to synchronize its clock with the edge-cloud, the two clocks are not guaranteed to be synchronized exactly. Due to possible clock offsets, timestamps from vehicle and cloud cannot be compared directly to compute 5G communication latencies. Instead, we assign the same unique identifier to each sample and its corresponding timestamps, such that the total round trip communication latency between the two \lstinline{mqtt_client} instances can be computed as the difference between the total latency and the sum of all other propagation and computation latencies.

Each experiment averages metrics over \num{600} point cloud transmissions, i.e., over a duration of \SI{60}{\second}, if run at \SI{10}{\hertz}. During each experiment, our research vehicle is standing still within \( \sim \SI{150}{\metre} \) distance of the next 5G antenna. The vehicle computer is equipped with two 10-core \textit{Intel~XEON~E5-2650V4} CPUs and an \textit{NVIDIA~GeForce~GTX~1080}. The server is equipped with two 64-core \textit{AMD~EPYC~7742} CPUs and four \textit{NVIDIA~A100} GPUs with \SI{40}{\giga\byte} memory each. All software components are executed in Docker containers running \textit{Ubuntu~20.04}, \textit{ROS~Noetic}, and \textit{TensorFlow~2.6}. All ROS components are implemented in C++ and launched as ROS nodelets to avoid memory copies between nodelets on the same machine.

As testbed, we use the 5G-Industry Campus Europe at RWTH Aachen University's Melaten campus~\cite{5G_Industry_Campus_Europe}. The 5G NSA network operates within a frequency band of \SIrange{3.7}{3.8}{\giga\hertz} and is accompanied by a \SI{2.3}{\giga\hertz} 4G anchor band. The edge-cloud server is connected to the university network, which the 5G network is directly tied to as well.

\subsection{Baselines}
\label{ssec:baselines}

In addition to the our custom MQTT-based approach to which we refer to as \textit{mqtt\_client} in the following, we perform experiments for three different baselines:
\begin{enumerate}
  \item \textit{mqtt\_bridge:} replacing our own C++ ROS-MQTT interface implementation with the publicly available ROS package \lstinline{mqtt_bridge}~\cite{mqtt_bridge}, which is based on Python;
  \item \textit{overlay:} connecting all involved ROS nodelets via an overlay network and using a single multi-host ROS master;
  \item \textit{in-vehicle:} running object detection inference locally on the vehicle computer, i.e., not relying on cloud connectivity at all.
\end{enumerate}

\textbf{mqtt\_bridge} --- The ROS Python package \lstinline{mqtt_bridge}~\cite{mqtt_bridge} can be integrated as a drop-in replacement for our own C++ ROS-MQTT interface. The data flow illustrated in \fig{fig:data-flow} remains unchanged.

\textbf{overlay} --- The direct connection between ROS nodelets on vehicle and edge-cloud can be achieved by launching only a single ROS master on one of the machines and connecting nodelets on the other machine to the same master via the \lstinline{ROS_MASTER_URI} environment variable. As we run all software in isolated Docker containers, we use a Docker swarm overlay network to connect the two containers among the two vehicle and cloud Docker daemon hosts. The data flow illustrated in \fig{fig:data-flow} changes in the way that point clouds published in-vehicle are directly transmitted to the object detection nodelet running in the edge-cloud and vice-versa for the computed object lists.

\textbf{in-vehicle} --- For the baseline case of local processing only, the published point cloud again is directly delivered to the object detection nodelet, this time running on the vehicle computer as well. In fact, all in-vehicle data transfers don't require any data copying. The object detection component in the data flow illustrated in \fig{fig:data-flow} moves to the vehicle side on the left, all other components are dropped.

\subsection{Experimental Variations}
\label{ssec:experimental-variations}

Apart from the baselines presented in \sect{ssec:baselines}, we also evaluate the impact of several experimental variations on the measured metrics:
\begin{enumerate}
  \item \textit{Encryption:} we enable encryption or use a VPN to secure the network traffic.
  \item \textit{QoS:} we test different MQTT QoS strategies;
  \item \textit{Data Type:} instead of sending messages of type \lstinline{VelodyneScan}, we convert to \lstinline{PointCloud2} before transmission;
  \item \textit{Data Size:} we evaluate the effect of data size by downsampling the transmitted point cloud.
\end{enumerate}

\textbf{Encryption} --- An important consideration in the context of connected driving is how to ensure a secure data transmission, which does not expose transmitted data to all potential listeners. The MQTT protocol supports exchanging data via SSL/TLS-encrypted communication channels. Similarly, all traffic via a Docker swarm overlay network can be AES-encrypted. As an alternative, we also consider connecting the vehicle to the university network via a VPN-tunnel.

\textbf{QoS} --- For the MQTT-based experiments, we test different Quality-of-Service~(QoS) levels. There are three QoS levels in MQTT, which formulate a guarantee of delivery of a specific message: the message is either delivered \textit{at most once (0)}, \textit{at least once (1)}, or \textit{exactly once (2)}.

\textbf{Data Type} --- The object detection nodelet expects \lstinline{PointCloud2} as input. Instead of sending the point cloud as \lstinline{VelodyneScan} and converting it on the cloud server, we can also transmit the more common format \lstinline{PointCloud2}. While the \textit{Velodyne} format is storing point cloud points as pairs of distance and azimuth angle values, the \lstinline{PointCloud2} format instead stores 3D-coordinates, thereby increasing the message size in our case from \SI{0.18}{\mega\byte} to \SI{1.08}{\mega\byte} per point cloud.

\textbf{Data Size} --- We also evalute the effect of decreasing the data size of the transmitted point cloud. To this end, the point cloud is randomly downsampled to \SI{75}{\percent}, \SI{50}{\percent}, \SI{25}{\percent}, and \SI{0}{\percent} of its original size before being published. The \lstinline{VelodyneScan} format is composed of many so-called packets, each of which presents an equidistant azimuth range of the entire \ang{360} sweep. Randomly removing certain packets therefore allows to downsample the point cloud to the desired ratio. \SI{0}{\percent} then relates to an empty point cloud (\num{0} point cloud packets), with the payload only consisting of a standard ROS message header of type \lstinline{std_msgs::Header}. Note that in all cases the same object list is sent back to the vehicle.

\subsection{Evaluation Metrics}
\label{ssec:evaluation-metrics}

For each point cloud sample published, the total round trip latency (in-vehicle ROS point cloud to in-vehicle ROS object list) as well as all partial latencies (cf.~points of timestamping in \fig{fig:data-flow}) are measured. This includes propagation latencies between components on the same machine, communication latencies between the two machines, and computation latencies of individual components.

Our main performance-indicating metric is the mean of all measured latencies of a single \SI{60}{\second} experiment. Additionally, we also compute minimum, maximum, and median latency as well as the corrected standard deviation~(jitter).

\section{Evaluation}
\label{sec:evaluation}

In this section, we compare the performance of our MQTT-based approach for 5G-enabled edge-cloud lidar object detection to the baselines presented in \sect{ssec:baselines}. We also evaluate the impact of multiple variations to the default experimental setup, as outlined in \sect{ssec:experimental-variations}.

Additionally, we analyze the results of experiments specifically designed to evaluate the capabilities of the 5G-network in terms of latency and throughput.

\subsection{Benchmarking of 5G Edge-Cloud Lidar Object Detection}
\label{ssec:benchmarking-of-5g-edge-cloud-lidar-object-detection}

The mean latencies measured for the proposed MQTT-based approach as well as the three baselines are reported in \tab{tab:methods-mean-latencies}. The total latency is split into partial contributions from different steps of the entire data flow, as illustrated in \fig{fig:data-flow}.

\begin{table}[!b]
  \centering
  \vspace{0.5\baselineskip}
  \setlength{\tabcolsep}{2pt}
  \newcommand*\circled[1]{\tikz[anchor=base, baseline]{\node[shape=circle, draw, inner sep=.2pt] (char) {#1};}}
  \newcommand*\circledf[1]{\tikz[anchor=base, baseline]{\node[shape=circle, fill, text=white, inner sep=.2pt] (char) {#1};}}
  \begin{tabular}{l l S S S S}
    \toprule
    & \textbf{Method} & \textbf{mqtt\_client} & \textbf{mqtt\_bridge} & \textbf{overlay} & \textbf{in-vehicle} \\
    \midrule
    \circledf{v}                   & \( \rightarrow \) MQTT interface   & 000.2 & 002.9 &       &       \\
    \circledf{v}                   & MQTT interface                     & 000.3 & 004.5 &       &       \\
    \hspace{3pt}\circled{\(\ast\)} & vehicle \( \rightarrow \) cloud    & 019.0 & 024.3 & 019.5 &       \\
    \hspace{6pt}\circledf{c}       & MQTT interface                     & 000.1 & 004.9 &       &       \\
    \hspace{6pt}\circledf{c}       & \( \rightarrow \) object detection & 003.6 & 002.7 &       & 003.5 \\
    \hspace{6pt}\circledf{c}       & object detection                   & 043.4 & 043.2 & 044.2 & 072.2 \\
    \hspace{6pt}\circledf{c}       & \( \rightarrow \) MQTT interface   & 000.1 & 000.7 &       &       \\
    \hspace{6pt}\circledf{c}       & MQTT interface                     & 000.1 & 002.3 &       &       \\
    \hspace{3pt}\circled{\(\ast\)} & cloud \( \rightarrow \) vehicle    & 019.0 & 024.3 & 019.5 &       \\
    \circledf{v}                   & MQTT interface                     & 000.1 & 006.2 &       &       \\
    \circledf{v}                   & \( \rightarrow \) benchmark IO     & 000.2 & 000.7 &       & 000.2 \\
    \midrule
                                   & \textbf{Total}                     & 086.1 & 116.7 & 083.2 & 075.9 \\
    \bottomrule
  \end{tabular}
  \caption[Mean latencies for different methods]{Mean latencies~[\si{\milli\second}] for different methods. \circledf{v}~indicates a vehicle latency component, \circledf{c}~indicates a cloud latency component. The communication latencies \circled{\(\ast\)} are averaged over both ways.}
  \label{tab:methods-mean-latencies}
\end{table}

\textbf{mqtt\_bridge} --- The direct comparison between our own C++- and the publicly available Python version of the ROS-MQTT interface reveals the superiority of C++: all partial latencies related to the interface are more than one magnitude faster than their Python counterparts, such that the total latency can be reduced from \SI{116.7}{\milli\second} to \SI{86.1}{\milli\second}. Apart from the obvious speed-up related to the interface implementation itself, message transmission to other ROS nodes on the same host is also faster most of the time, since no-copy message transfers between ROS C++ nodelets are possible.

\textbf{in-vehicle} --- The lowest mean total latency is achieved with in-vehicle processing: object lists are available on average \SI{75.9}{\milli\second} after a point cloud scan has taken place. This latency, however, is entirely dominated by the performance of the object detection algorithm on the local vehicle hardware. In fact, our edge-cloud runs are only \( \sim \SI{10}{\percent} \) slower than local processing thanks to an object detection runtime decrease by \( \sim \SI{40}{\percent} \).

\textbf{overlay} --- Compared to the single-ROS-master overlay network approach, our MQTT-based approach is slightly outperformed in terms of mean total latency (\SI{83.2}{\milli\second} vs.\ \SI{86.1}{\milli\second}), but brings advantages discussed further below. Although the baseline's summed communication latency between vehicle and cloud seems \SI{1}{\milli\second} slower, here it also contains the conversion of point cloud packets to the \lstinline{PointCloud2} format, which we have not measured separately. Based on the propagation latency between interface and object detection in the MQTT-based runs, we can estimate the point cloud conversion latency to \( \sim \SI{3}{\milli\second} \). Overall, both approaches perform similarly and are mostly bound by the performances of their common components: the actual data transmission (communication latency) and the object detection algorithm.

\begin{figure}[!t]
  \centering
  \includegraphics[width=\columnwidth]{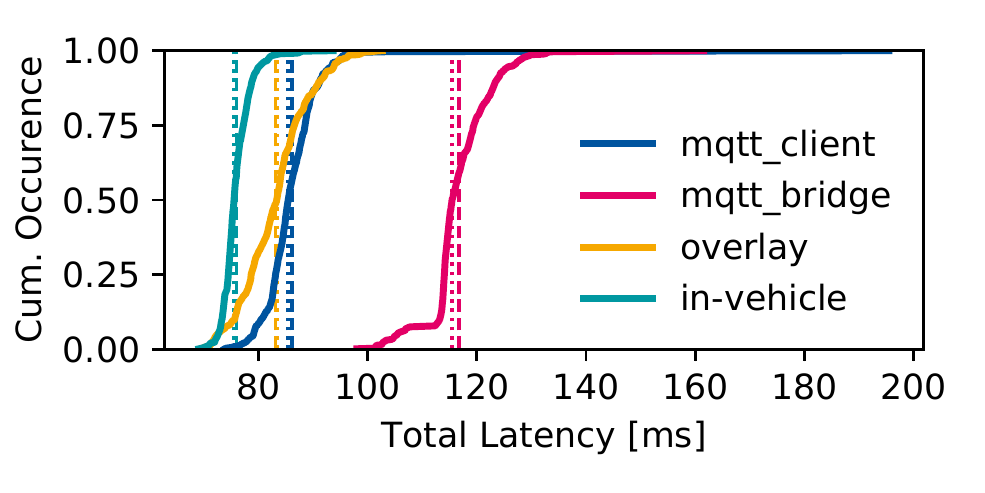}
  \caption{ECDF of total latency measurements for different methods. Means are included as dashed lines, medians as dotted lines.}
  \label{fig:methods-cumulative}
\end{figure}

In addition to the partial latencies given in \tab{tab:methods-mean-latencies}, the empirical cumulative distribution function~(ECDF) of total latency measurements is plotted in \fig{fig:methods-cumulative}. The distribution shows that the in-vehicle approach is associated with the lowest and the Python MQTT interface comes with the highest variance in total latency. For our \lstinline{mqtt_client} and for the overlay network approach, we measure similar standard deviations of total latency: \SI{6.7}{\milli\second} and \SI{6.1}{\milli\second}, respectively. It is also worth noting that in both these cases \SI{99.5}{\percent} of the \num{600}~total latencies are below \SI{100}{\milli\second}.

Even though our proposed MQTT interface incurs a little overhead compared to the overlay network solution, it successfully separates local coordinator functions in vehicle and cloud from communication-coordinating functions, i.e., the MQTT broker. Computation and communication in the overlay solution is entirely dependent on the one ROS master process, thus posing a single point of failure. The single master process also makes scaling to many intelligent agents a lot harder, whereas the separate publish-subscribe framework of MQTT is well suited for e.g.\ transmitting data from many agents to one edge-cloud server. Finally, the MQTT-based approach can also integrate with non-ROS clients running on, e.g., constrained IoT devices or microcontrollers.

\vspace{0.25cm}
\textbf{Encryption} --- \tab{tab:encryption-mean-latencies} contains measured mean latencies for experiments, where encrypted communication channels have been established to secure the data transmission. The introduced overhead is visible in the partial latencies of the MQTT interface processing and the communication latencies, any other partial latencies should not be affected. 

While the sum of the \lstinline{mqtt_client} latencies increases by \( \sim \SI{67}{\percent} \) due to en- and decryption, it still only makes up \( \sim \SI{1}{\percent} \) of the total latency. Whether the equally small increase in communication latency can be attributed to encryption, cannot be answered with certainty. The effect of encryption is more noticeable in the overlay approach, where communication latencies increase by \( \sim \SI{20}{\percent} \).

Using a VPN-tunnel to connect to the MQTT broker introduces a similarly large overhead on communication latency as the overlay network encryption.

\begin{table}[!hb]
  \centering
  \begin{tabular}{l S S S S S}
    \toprule
    \textbf{Method} & \multicolumn{3}{c}{\textbf{mqtt\_client}} & \multicolumn{2}{c}{\textbf{overlay}} \\
    \textbf{Encryption} & {\textbf{none}} & {\textbf{SSL}} & {\textbf{VPN}} & {\textbf{none}} & {\textbf{AES}} \\
    \midrule
    MQTT interfaces                     & 000.6 & 001.0 & 001.0 &       &       \\
    vehicle \( \leftrightarrow \) cloud & 038.0 & 039.0 & 044.4 & 038.9 & 046.6 \\
    \midrule
    \( \Sigma \)                        & 038.6 & 040.0 & 045.4 & 038.9 & 046.6 \\
    \bottomrule
  \end{tabular}
  \caption{Effect of encryption on relevant mean latencies~[\si{\milli\second}].}
  \label{tab:encryption-mean-latencies}
\end{table}

\textbf{QoS} --- \tab{tab:qos-mean-latencies} reports measurements for experiments with \lstinline{mqtt_client} at varying MQTT QoS levels. One thing to note is that a significant part of the communication latencies moves to the measurements of MQTT interface latencies for QoS values greater than 0. This effect is due to the implemented timestamping mechanism in \lstinline{mqtt_client}: the final timestamp is not taken until the send-function has returned, which in the case of higher QoS values has to wait for a message reception acknowledgement from the broker.

For QoS 1, only a little overhead of \( \sim \SI{5}{\percent} \) in the sum of interface and communication latencies is observed. A QoS value of 2 however --- guaranteeing that the messages are delivered exactly once --- more than doubles the transmission latency to \SI{93.1}{\milli\second}. The latency increase with higher QoS levels is expected, since a single message transmission then comprises one (QoS~1) or two request/response flows (QoS~2). Also note that the guarantee of delivery is enforced for the vehicle-broker-, broker-cloud-, cloud-broker-, and broker-vehicle-transmissions.

\begin{table}[!hb]
  \centering
  \begin{tabular}{l S S S}
    \toprule
    \textbf{Method} & \multicolumn{3}{c}{\textbf{mqtt\_client}} \\
    \textbf{QoS} & {\textbf{0}} & {\textbf{1}} & {\textbf{2}} \\
    \midrule
    MQTT interfaces                     & 000.6 & 038.2 & 058.3 \\
    vehicle \( \leftrightarrow \) cloud & 038.0 & 002.2 & 034.8 \\
    \midrule
    \( \Sigma \)                        & 038.6 & 040.4 & 093.1 \\
    \bottomrule
  \end{tabular}
  \caption{Effect of MQTT QoS level on relevant mean latencies~[\si{\milli\second}]}
  \label{tab:qos-mean-latencies}
\end{table}

\textbf{Data Size and Type} --- \tab{tab:data-size-mean-latencies} shows how the measured latencies depend on data size. For interface processing and object detection latencies, a near-linear trend can be observed. Any differences in communication latencies are however still within the bounds of network noise. Since the communication and object detection latencies are dominated by a constant overhead, irrelevant of data size, the total latency of the process can only be decreased by \( \sim \SI{13}{\percent} \), even when discarding \( \sim \SI{75}{\percent} \) of the original point cloud data.

At \SI{0}{\percent} point cloud size, the message payload only consists of a header. Still, the summed communication latency incl. MQTT interface only falls to \( \sim \SI{74}{\percent} \) of that of the full point cloud reference case. Note that the full object list is sent back to the vehicle though.

One variation related to data size is transmitting \lstinline{PointCloud2} messages instead of point cloud packets. The effect of a 6-times larger payload size is clearly visible in both MQTT interface as well as communication latencies.

\begin{table}[!hb]
  \centering
  \setlength{\tabcolsep}{3pt}
  \begin{tabular}{l S S S S S S}
    \toprule
    \textbf{Method} & \multicolumn{6}{c}{\textbf{mqtt\_client}} \\
    \textbf{Type} & \multicolumn{5}{c}{\textbf{VelodyneScan}} & {\textbf{PointCloud2}} \\
    \textbf{Size} & {\textbf{0\%\tablefootnote{This experiment was conducted with a MQTT QoS level of 1 to avoid oscillations that occured at QoS 0.}}} & {\textbf{25\%}} & {\textbf{50\%}} & {\textbf{75\%}} & {\textbf{100\%}} & {\textbf{100\%}} \\
    \midrule
    MQTT inferfaces                     & 026.8 & 000.3 & 000.4 & 000.5 & 000.6 & 006.1 \\
    vehicle \( \leftrightarrow \) cloud & 001.8 & 034.6 & 033.0 & 033.0 & 038.0 & 140.4 \\
    object detection                    & 000.0 & 038.9 & 040.3 & 042.8 & 043.4 & 047.6 \\
    \midrule
    \( \Sigma \)                        & 028.6 & 073.8 & 073.7 & 076.3 & 082.0 & 194.1 \\
    \bottomrule
  \end{tabular}
  \caption{Effect of data size on relevant mean latencies~[\si{\milli\second}].}
  \label{tab:data-size-mean-latencies}
\end{table}

\subsection{Additional Benchmarking of 5G Network Performance}
\label{ssec:additional-benchmarking-of-5g-network-performance}

In order to separately evaluate the potential of the MQTT-based connection via 5G, unrelated to the presented use case of edge-cloud lidar object detection, we also report one-way communication latency measurements for transmission-only experiments. To this end, we measure round trip times of sending point clouds from the vehicle to the MQTT broker and back, while averaging communication latencies.

\begin{figure}[!htbp]
  \centering
  \includegraphics[width=\columnwidth]{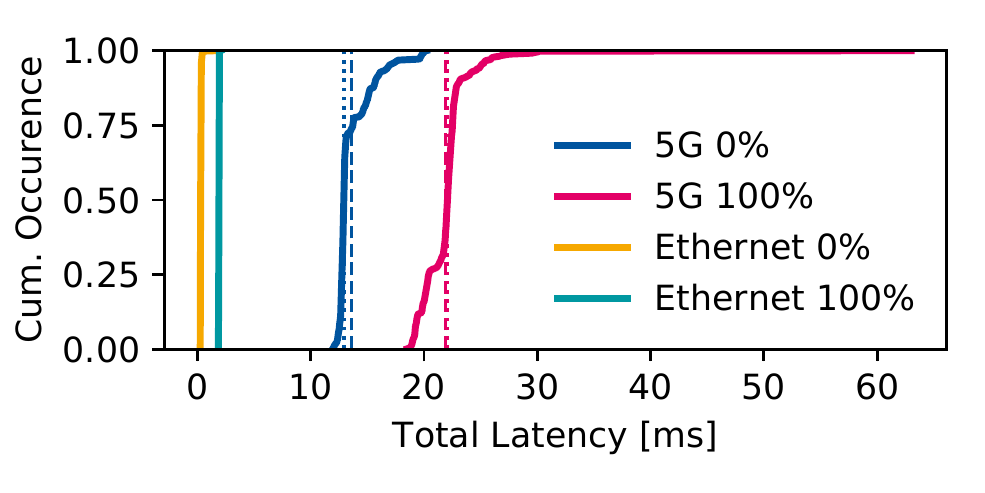}
  \caption{ECDF of one-way communication latency measurements via Ethernet and 5G. Means are included as dashed lines, medians as dotted lines.}
  \label{fig:flighttime-cumulative}
\end{figure}

\tab{tab:flighttime-metrics} includes one-way 5G communication latency metrics compared to a physical Ethernet connection. The ECDF of all transmitted samples is additionally plotted in \fig{fig:flighttime-cumulative}. Communication via 5G is on average more than 10~times slower than transmission via Ethernet. Still, a one-way communication latency as low as \SI{18.5}{\milli\second} for a full point cloud sample can be reached via 5G. The displayed distribution however also highlights that 5G communication is subject to a lot more variance as compared to Ethernet.

\begin{table}[!htbp]
  \centering
  \begin{tabular}{l S S S S}
    \toprule
    \textbf{Method} & \multicolumn{4}{c}{\textbf{mqtt\_client} (communication-only)} \\
    \textbf{Network} & \multicolumn{2}{c}{\textbf{5G}} & \multicolumn{2}{c}{\textbf{Ethernet}} \\
    \textbf{Size} & {\textbf{0\%}} & {\textbf{100\%}} & {\textbf{0\%}} & {\textbf{100\%}} \\
    \midrule
    Mean   & 013.6 & 021.9 & 000.3 & 001.9 \\
    Median & 012.9 & 022.1 & 000.3 & 001.9 \\
    Min    & 011.9 & 018.5 & 000.2 & 001.8 \\
    Max    & 020.2 & 063.0 & 001.6 & 002.0 \\
    Std    & 001.6 & 002.4 & 000.1 & 000.0 \\
    \bottomrule
  \end{tabular}
  \caption{One-way communication latency metrics~[\si{\milli\second}] via 5G and Ethernet.}
  \label{tab:flighttime-metrics}
\end{table}

We also try to push the 5G connection to its limits by increasing publishing rates beyond \SI{10}{\hertz}. To this end, the same setup as illustrated in \fig{fig:data-flow} is used, except that the object detection component is replaced with a simple loop-through of the point cloud, which is also being sent back to the vehicle.

At a publishing rate of \SI{55}{\hertz}, the maximum sustainable throughput with a symmetric up- and downlink of \( \sim \SI{10}{\mega\byte\per\second} \) is reached. Communication becomes unstable at \SI{56}{\hertz} (\SI{10.27}{\mega\byte\per\second}) in the sense that latencies fluctuate heavily. The 5G transmission is verifiably identified as the bottleneck at this data rate, since no point cloud sample is being dropped, despite restricting the ROS-MQTT interface to process only one sample at a time. Note that the limiting factor here is up- and not downlink. It is also noteworthy that mean latencies show no clear dependence on data rate up to the critical point.

\section{Conclusion}
\label{sec:conclusion}

We have proposed a novel MQTT-based universal communication interface for intelligent robotic agents running ROS. The interface has been evaluated in an extensive 5G case study in the context of connected automated driving, involving edge-cloud object detection in lidar point clouds. We have compared our interface against multiple alternative approaches, highlighting the on-par latency of our solution compared to the best baseline, while at the same time bringing more flexibility, reliability, and scalability. Additionally, we have analyzed latencies under multiple experimental variations such as encryption and data size.

The developed ROS-MQTT interface as well as all tooling and data involved in benchmarking is released open-source and free to use. Step-by-step guides enable anyone to analyze the presented use case in their own network. Furthermore, the released \lstinline{mqtt_client} ROS package contributes not only to automated driving applications, but also to the robotics and IoT ecosystems as a whole.

The work at hand motivates further research in both mobile networking technologies as well as other potential use cases in the field of connected driving and beyond. The released tools enable benchmarking comparisons between 4G, 5G, and future 6G communication technology. Based on realizable latencies, bandwidth, and availability, offloading entire driving functions to edge-clouds or remote operators becomes imaginable.


\section*{Acknowledgement}

This research is accomplished within the projects 6GEM (FKZ~16KISK036K) and UNICAR\textit{agil} (FKZ~16EMO0284K).
We acknowledge the financial support for the projects by the Federal Ministry of Education and Research of Germany~(BMBF).

\bibliographystyle{IEEEtran}
\bibliography{references}

\end{document}